\documentclass[letterpaper, 10 pt, conference]{ieeeconf}  

\IEEEoverridecommandlockouts                              
\usepackage{graphicx}
\usepackage{amsmath}
\usepackage{amssymb}
\usepackage{csquotes}
\usepackage{booktabs}
\usepackage{soul}
\usepackage{amsmath}
\usepackage{paracol}
\usepackage{empheq}
\usepackage[table]{xcolor}
\usepackage{tabularx}
\usepackage{multirow}
\usepackage[pagebackref=true,breaklinks=true,letterpaper=true,colorlinks,bookmarks=false]{hyperref}
\usepackage{paracol}

\usepackage{xspace}
\usepackage[capitalize]{cleveref}
\crefname{section}{Sec.}{Secs.}
\Crefname{section}{Section}{Sections}
\Crefname{table}{Table}{Tables}
\crefname{table}{Tab.}{Tabs.}
\usepackage{tikz,xcolor}

\definecolor{lime}{HTML}{A6CE39}
\DeclareRobustCommand{\orcidicon}{%
    \begin{tikzpicture}
    \draw[lime, fill=lime] (0,0) 
    circle [radius=0.16] 
    node[white] {{\fontfamily{qag}\selectfont \tiny ID}};
    \draw[white, fill=white] (-0.0625,0.095) 
    circle [radius=0.007];
    \end{tikzpicture}
    \hspace{-2mm}
}
\newcommand{\latinphrase}[1]{\textit{#1}}
\newcommand{\etal}{\latinphrase{et~al.}\xspace}

\newcommand{\etc}{\latinphrase{etc.}\xspace}

\newcommand{\ibrahim}{\href{https://orcid.org/0000-0002-5376-2477}{\orcidicon}}
\newcommand{\naveed}{\href{https://orcid.org/0000-0003-3406-673X}{\orcidicon}}
\newcommand{\saeed}{\href{https://orcid.org/0000-0002-0692-8411}{\orcidicon}}
\newcommand{\ajmal}{\href{https://orcid.org/0000-0002-5206-3842}{\orcidicon}}

\title{\LARGE \bf
UnLoc: A Universal Localization Method for Autonomous Vehicles\\ using LiDAR, Radar and/or Camera Input
}

\author{Muhammad Ibrahim\ibrahim$^{1}$,  Naveed Akhtar\naveed$^{1}$,  Saeed Anwar\saeed$^{2}$, and  Ajmal Mian\ajmal$^{1}$ 
\thanks{Professor  Ajmal  Mian  is  the  recipient  of  an  Australian Research Council Future Fellowship Award (project number FT210100268) funded by the Australian Government.
Dr.~Naveed Akhtar is recipient of an Office of National Intelligence National Intelligence Postdoctoral Grant (project number NIPG-2021-001)
funded by the Australian Government.}% <-this % stops a space
\thanks{$^{1}$ Department of Computer Science, The University of Western Australia.
        {\tt\small muhammad.ibrahim@research, naveed.akhtar@, ajmal.mian@) uwa.edu.au}
        }%
\thanks{$^{2}$Information and Computer Scicence, King Fahad University of Petroleum and Minerals (KFUPM), Dhahran, KSA, %Lecturer and Visiting Fellow at University of Technology Sydney. 
        {\tt\small saeed.anwar@kfupm.edu.sa}}%
}

\makeatletter
\def\endthebibliography{%
  \def\@noitemerr{\@latex@warning{Empty `thebibliography' environment}}%
  \endlist
}
\makeatother
\begin{document}

\maketitle
\thispagestyle{empty}
\pagestyle{empty}

%%%%%%%%%%%%%%%%%%%%%%%%%%%%%%%%%%%%%%%%%%%%%%%%%%%%%%%%%%%%%%%%%%%%%%%%%%%%%%%%
\begin{abstract}
Localization is a fundamental task in robotics for autonomous navigation. Existing localization methods rely on a single input data modality or train several computational models to process different modalities. This leads to stringent computational requirements and sub-optimal results that fail to capitalize on the complementary information in other data streams. This paper proposes UnLoc, a novel unified neural modeling approach for localization with multi-sensor input in all weather conditions. Our multi-stream network can handle LiDAR, Camera and RADAR inputs for localization on demand, i.e., it can work with one or more input sensors, making it robust to sensor failure. UnLoc uses 3D sparse convolutions and cylindrical partitioning of the space to process LiDAR frames and implements ResNet blocks with a slot attention-based feature filtering module for the Radar and image modalities. We introduce a unique learnable modality encoding scheme to distinguish between the input sensor data. Our method is extensively evaluated on Oxford Radar RobotCar, ApolloSouthBay and Perth-WA datasets. The results ascertain the efficacy of our technique.
%Localization is a fundamental task in robotics for autonomous navigation. Existing localization methods either rely on a single input data modality, or train different  computational models to process different  modalities. This not only leads to stringent computational requirements, but also sub-optimal results that fail to capitalize on the complementary information in different data streams. In this paper, we propose UnLoc, a novel unified neural modeling approach for localization with  multi-sensor input in all weather conditions. Our multi-stream network can handle LiDAR, Camera and RADAR inputs for localization on demand i.e., it can work with one or more input sensors which also also makes it robust to sensor failure. UnLoc uses 3D sparse convolutions and cylindrical partitioning of the space to process LiDAR frames, and implements ResNet blocks with a slot attention-based feature filtering module for the Radar and image modalities. We introduce a unique  learnable modality encoding scheme to distinguish between the input sensor data. Our method is extensively  evaluated on Oxford Radar RobotCar, ApolloSouthBay and Perth-WA datasets. The results ascertain the efficacy of our technique. 
\end{abstract}

%%%%%%%%%%%%%%%%%%%%%%%%%%%%%%%%%%%%%%%%%%%%%%%%%%%%%%%%%%%%%%%%%%%%%%%%%%%%%%%%
\section{INTRODUCTION}
\vspace{-1.3mm}
Vehicle localization in outdoor environment is an essential task in robotics, especially in the autonomous driving domain. To achieve self-autonomy in urban outdoor environment, a vehicle must be able to precisely localize itself. Current outdoor localization  systems rely on the Global Navigation Satellite System (GNSS). However, the lack of accuracy and signal blockage in densely populated regions for GNNS make it an inadequate technology for autonomous vehicles.
Creating an offline map of the environment and using query frames during online navigation provides a viable alternate solution to the problem. Conventional methods in this direction~\cite{segal2009generalized, stoyanov2012fast} employ frame registration for localization. However, this leads to  impractical computational requirements.  More  recently, deep learning techniques have shown great promise in addressing the issue~\cite{ibrahim2023slice}.

Among the deep learning methods, 3D point cloud regression-based approaches, e.g., \cite{ibrahim2023slice, wang2021pointloc}, can directly predict six degrees of freedom (6DoF) poses to localize vehicles.
%These metheds exploiting PointNet and Transformer~\cite{vaswani2017attention}, respectively. 
Point clouds provide depth information of the scene and $360^{\circ}$ field of view (FoV), which are helpful for precise localization. However, LiDAR data is also inherently prone to high level of noise in rainy and foggy weather. Moreover, its unstructured nature demands complex and  computationally expensive modeling when outdoor localization  completely relies on the LiDAR data. 
%due to the unstructured nature of the point cloud data, it demands complex representation learning, which adds to the computational requirements of the model. 

%3D point cloud regression-based approaches~\cite{wang2021pointloc}, \cite{ibrahim2023slice} are proposed which directly predicts six degrees of freedom (6DoF) poses to localize vehicles over the map exploiting pointnet and transformer~\cite{vaswani2017attention} respectively. Point cloud modality provides depth information and $360^{\circ}$ field of view (FoV) which allows the localization process more practical. However, LiDAR sensor generates noisy data in rainy and foggy weather due to defection of light rays. Also, due to unstructured nature of the data, it demands high-fidelity representation learning by the model. 

In the related literature, processing RGB camera images with deep learning  is also  considered suitable for localization.  
%Another important deep-learning localization is based on image modality. 
Currently, techniques such as PoseNet and its variants~\cite{kendall2015posenet, kendall2016modelling} use  a single or a series of image frames to predict 6DoF poses. Whereas image modality offers detailed spatial information, it is easily influenced by  environmental variations, such as sunlight, rain, fog etc., which is detrimental for localization. Comparatively, Radar data  is largely insensitive to various weather conditions, e.g., darkness, fog, snow and sunlight. Leveraging that,  Cen~\etal~\cite{cen2018precise} extracted features from Radar scans and then applied scan matching to predict ego-motion. Radarloc~\cite{wang2021radarloc} is a recent deep learning localization method that predicts global poses from Radar data.
Nevertheless, Radar data does not have precise 3D information and is noisy, which compromises the overall localization performance. 

\begin{figure*}
\begin{center}
\includegraphics[width=1\textwidth, height=0.29\textheight]{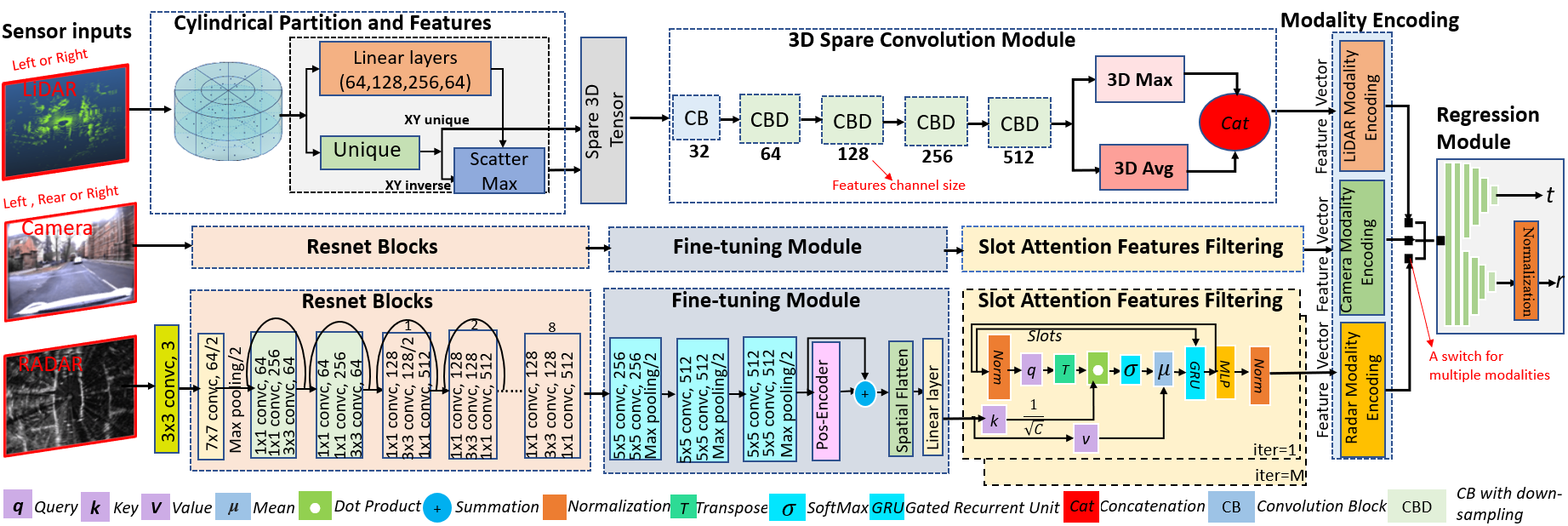}
\end{center}
\vspace{-3mm}

\caption{Architecture of the proposed UnLoc. 
\textbf{Top} stream of layers of our network processes 3D point cloud data. It transforms the input into a cylindrical representation which is processed by sparse 3D convolution blocks,  CB and CBD (see Fig.~\ref{fig:cbd}), to extract point cloud features. These features are passed through 3DMax and 3DAvg layers to compute the feature vectors. \textbf{Middle} stream of the network processes images to extract features with ResNet blocks, which are further processed by our fine tuning and a transformer based features filtering module. \textbf{Bottom} stream processes Radar modality and has the same architecture as the middle stream, except with  an additional 2D Conv layer. Feature vectors for each modality are encoded by our  learnable modality encoding scheme which passes the features to a Regression Module for 6DoF poses prediction.}
%\textcolor{red}{See comments in latex}}%NA: I have made a few minor changes in the section names. Make sure that the block names in this figure follow titles of the related subections. Also, try to make the text in the leged (at the very bottom) a little larger. Ib: Done 
\label{fig:method}
\vspace{-5mm}
\end{figure*}

For the applications like self-driving vehicles, robust outdoor localization is only possible by leveraging complementary characteristics of different data modalities.   
%For self-driving vehicles, localization in outdoor environments requires a multi-sensors deep learning architecture to achieve high accuracy in all weather conditions. 
In this work, we present a multi-sensor localization approach, shown in Fig.~\ref{fig:method}, that learns a unified neural model, called UnLoc,  for point cloud, Radar and image data, to achieve  precise 6DoF localization. The proposed model  uses sparse 3D convolutions to process cylindrical representations of point clouds, while 2D convolutions and slot attention-based feature filtering are used to process Radar and image modalities. We also introduce a learnable modality encoding technique to optimally discriminate between different data modalities for their on demand use. Our method allows the use of a single or any combination of modalities during inference. This makes our method robust to sensor failure.

Our network is trained on six sensor inputs for  three different modalities. Due to the unique universal nature of our method, we also devise a technique to generate common ground-truth for different sensor data streams, which enables effective training of our model.  
%At inference time, either a single sensor or multiple sensor data can be used to predict the final pose. 
Our method  is an adaptive deep-learning technique that can be used in challenging real-world environments.
We establish the performance of our approach on three publicly available datasets:  Oxford Radar RobotCar~\cite{RadarRobotCarDatasetICRA2020}, Appolo-Southbay~\cite{L3NET_2019_CVPR}, and Perth-WA~\cite{ibrahim2023slice}. 
%At first, we setup our model for Oxford Radar RobotCar~\cite{RadarRobotCarDatasetICRA2020}, which is later fine-tuned for the other two datasets. 
We also conduct a thorough ablation study to analyze the effects of using various sensor inputs with our model. We establish strong localization  results on all the datasets, outperforming the existing methods on each modality and further improving the performance by processing multi-modality input.

\section{Related Work}
\vspace{-1mm}
Localization is essential for autonomous driving~\cite{elhousni2020survey, gao2021fully}. In the existing literature, different input data modalities, such as point cloud, image, and Radar data, distinguish different   localization methods. 
Conventional point cloud matching techniques employ registration methods~\cite{segal2009generalized, kovalenko2019sensor}. On the other hand, recent point cloud deep learning-based techniques associate input frames to a point cloud map for 6DoF pose estimation~\cite{wang2021pointloc, L3NET_2019_CVPR, nubert2021self, dube2018segmap}. Among the mentioned methods,  PointLoc~\cite{wang2021pointloc}, Slice~\cite{ibrahim2023slice} and L3Net~\cite{L3NET_2019_CVPR} compress the map into a neural 6DoF pose predictor for vehicles. The PointLoc~\cite{wang2021pointloc} exploits PointNet framework to predict poses while Slice~\cite{ibrahim2023slice} uses transformer architecture~\cite{vaswani2017attention}. In general, directly predicting 6DoF pose from a point cloud frame is challenging because the  unstructured  LiDAR   representation conflicts with  the high precision demands of the task. Therefore, other methods, e.g., ~\cite{kendall2015posenet, kendall2016modelling} often  augment their neural models to handle the data complexity.

Some works also devise camera-based deep learning localization methods.  For instance, PoseNet~\cite{kendall2015posenet} is the pioneering technique that utilizes camera images to predict 6DoF poses. Likewise, the recent variants~\cite{kendall2016modelling, brahmbhatt2018geometry} of PoseNet regress poses using a single or multiple images, exploiting geometric loss and modeling poses with Bayesian Neural Network (BNN) to enhance the performance. Walch~\etal~\cite{walch2017image} employed LSTMs for matching geometric features to improve the pose precision. Retrieval-based learning approaches, e.g.,  CamNet~\cite{ding2019camnet}, RelocNet~\cite{balntas2018relocnet}, and Camera Relocalization CNN~\cite{laskar2017camera},  use agents that have previously visited the exact location. However, the above approaches are restricted in performance due to the demerits of the visual sensors.

% Recently, camera-based deep learning localization methods have been proposed that directly predict 6DoFs. PoseNet~\cite{kendall2015posenet} is the initial deep-learning localization method that utilizes camera images to predict 6DoF poses directly. The updated and improved versions of PoseNet are~\cite{kendall2016modelling, brahmbhatt2018geometry} that use single or multiple images to directly regress poses. They exploit geometric loss and modeling poses with Bayesian NN to enhance the performance.  Walch~\etal~~\cite{walch2017image} utilize LSTM architecture for geometric features matching to improve the precision of the pose. There are some retrieval-based learning approaches like CamNet~\cite{ding2019camnet}, RelocNet~\cite{balntas2018relocnet}, and Camera Relocalization CNN~\cite{laskar2017camera} in this category which require agents that have previously been visited the same place. Though these methods provide better regressions, they are still restricted to the demerits of visual sensors.

Contemporary localization strategies also explore the Radar data modality. For example, Barnes~\etal~\cite{barnes2019masking} proposed a deep correlative scan matching technique based on learned feature embedding and a self-supervised module for Radar odometry system. Later, the authors developed a deep key point detector and metric localizer~\cite{barnes2020under} for Radar odometry estimation. Cen~\etal~\cite{cen2018precise} extracted features from Radar scans and then applied scan matching to predict ego motion. RadarLoc~\cite{wang2021radarloc} is the latest method that predicts global absolute poses with respect to the world coordinate system employing Radar data. Compared to camera and LiDAR, radar is not as sensitive to the weather conditions and %independent, allowing it to operate in all weather conditions, e.g., darkness, fog, and snow. 
provides  $360^{\circ}$ FoV of a scene. % which is an advantage over the visual sensor. 
However, it lacks precise 3D information when compared to the LiDAR data.

%Radar-based modality is explored by recent localization methods. Barnes~\etal~~\cite{barnes2019masking} proposed a deep correlative scan matching technique based on learned feature embedding and self-supervised module for radar odometry system. Later, they also developed a deep key point detection and metric localization method~\cite{barnes2020under} for radar odometry estimation. Cenet al.~\cite{cen2018precise} extracted features from radar scans and then applied scan matching to predict ego. Radarloc~\cite{wang2021radarloc} is the latest deep learning localization method that predicts global absolute poses concerning the world coordinate systems employing radar images.  Radar modality is weather independent which allows it to operate at all weather conditions e.g. darkness, fog, and snow. Though it provides  $360^{\circ}$ FoV of a scene which is an advantage over visual sensor, but it lacks the depth information as compared to LiDAR sensor 

The methods mentioned above mainly rely on a  single sensor data modality. However, outdoor localization, especially for  autonomous driving, requires the precision and robustness that is hard to achieve with a single  modality.
%type since individual sensor data harbors its pros and cons. 
Still, localization through multiple sensors (Radar, camera and LiDAR) in  outdoor environments is currently a largely   unexplored direction. In this work, we fill this gap by devising a universal localization neural model that leverages point cloud, Radar and/or camera input on demand, to provide robust and precise 6DoF pose predictions.   

\label{sec:Method}
\section{METHODOLOGY}
We propose a multi-sensors localization method that can leverage point cloud, Radar and/or image modalities on demand. Our approach uses three parallel streams for these modalities in the early stages of the model, see Fig.~\ref{fig:method}. It applies sparse 3D convolutions on the LiDAR data and employs ResNet blocks followed by a slot attention-based feature filtering for the Radar and image modalities. Moreover, it employs  a learnable modality encoding that learns to identify the input sensor data stream for optimal performance. The architecture for the Radar and image data streams are largely  similar, except for  an additional 2D convolution layer to process the Radar data. Our technique computes a feature vector for each data modality and applies modality encoding to that. The individual data streams get projected onto their respective feature spaces that have similar dimensionality across modalities. The data  features eventually get processes by a regression module to predict the 6DoF pose. 
%size of the features vector for all modalities is the same, i.e., 1024. 
%Finally, it uses a standard regression module for all modalities to predict the 6DoF ground truths.
Our model is trained on six sensor inputs: LiDAR (left and right), camera (left, rear and right), and Radar. For inference, it accepts any single input or any combination of the input modalities. Our method is designed for localization in outdoor  environment, particularly for self-driving vehicles. Due to its multi-modality nature, it is well-suited to different weather conditions and is robust to sensor failures. Each component of our framework is explained in detail below.    

\label{sec:3dfeatures}
\subsection{3D Features Extraction}
We propose a sparse 3D convolution localization block to process the  point cloud data, see Fig.~\ref{fig:method}. This block  extracts 3D geometric features from LiDAR data, which are  particularly relevant for localization in the outdoor environment. The 3D convolution using voxelization is known for it efficacy to process  LiDAR data~\cite{zhu2020cylindrical}. However, voxel processing using 3D convolution is computationally expensive. Hence, we devise a lightweight sparse 3D convolutional method utilizing cylindrical partitioning. Our approach uses  a series of sparse 3D convolutional sub-blocks for processing point cloud data, followed by Max-pooling and Average pooling layers.  Details of this process are provided below.  
%This block is first trained on the Perth-WA point cloud dataset before plugging it into our framework.

\begin{figure}
\begin{center}
\includegraphics[width=0.8\columnwidth]{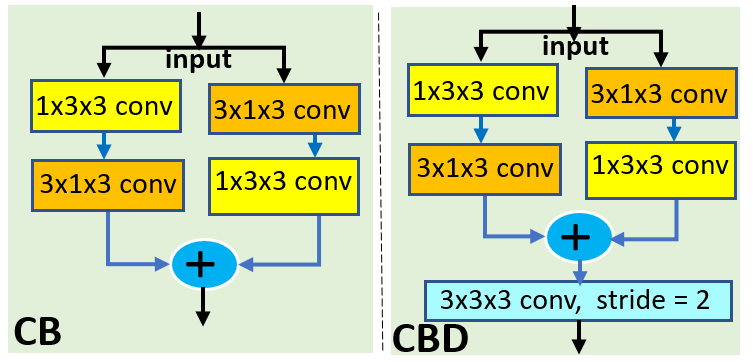}
\caption{Sparse 3D convolution block without down-sampling (\textbf{CB}) and with down-sampling (\textbf{CBD}, using an extra 3D convolution with stride 2).  \vspace{-7mm}}
\label{fig:cbd}
\end{center}
\vspace{-4mm}

\end{figure}

\subsubsection{3D Cylindrical Partition and Features}
Outdoor  LiDAR point cloud has the attribute of changing density, with nearby regions  having substantially higher point densities than the far regions. Hence, cylindrical coordinate system is well-suited to  partition LiDAR data,   
%we replace the Cartesian grid partition is replaced by the cylinder coordinate system in this block. 
which evenly distributes the points across various partitions by providing larger volumes for the far-off points. 
%and aligns with the distribution of outdoor points by placing them in the farther-off region. 
%Furthermore, it also helps preserving the geometric structure, unlike these projection-based approaches that project the point to the 2D image. 
The top-left corner in Fig.~\ref{fig:method} illustrates  the workflow of our partitioning, followed by point feature processing. First, the  $(X, Y, Z)$ coordinates of points  are converted to $(R,\theta,$ and $z)$ for a cylindrical grid representation, where $R$, $\theta$ are the radius and height, respectively. Then, a cylindrical partitioning is performed to generate voxels having largely uniform point distribution. The farther regions have larger voxels compared to the nearby regions. Next, the cylindrical grid representation is passed through an MLP-based module with linear layers to obtain cylinder features. These features are further processed %to obtain the maximum magnitude of cylinder features 
using Unique indices and Scatter-Max layers to obtain maximum magnitude features. Finally, we get the 3D cylinder representations with size  $\in \mathbb{R}^{D \times H \times W \times L}$, where $D$ is the feature dimension, and $H, W, L$ respectively represent the height, width, and length of the cylinder. Our 3D Sparse  convolution module subsequently  processes this representation. 

\subsubsection{3D Sparse Convolution}
To process the cylindrical representation, we had two options: conventional 3D convolution or sparse 3D convolution. We choose the latter for memory and computational efficiency. Inspired by Cylinder3D~\cite{zhu2020cylindrical}, we create two asymmetric types of 3D sparse convolution modules, without down-sampling (CB) and with down-sampling (CBD), as shown in Fig.~\ref{fig:cbd}. Both convolution blocks have two sparse convolution streams with stride 1. The first stream has kernel $(1,3,3)$ followed by kernel $(3,1,3)$ and the second stream has the same kernels in the reverse order. The output of both streams is added. In the convolution with down-sampling (CBD), an additional 3D convolution with kernel $(3,3,3)$ and stride 2 is applied for downsampling.
%The Convolution Block with Down-sampling (CBD) uses a sparse 3D convolution with stride 2. The  layers in the CBD  modules have kernel sizes $(1,3,3)$ and $(3,1,3)$, except for the last layer, which has a kernel size $(3, 3, 3)$ with stride 2.  
%The downsampling which has approximately 33\% higher computational cost~\cite{zhu2020cylindrical}.
 We employ  one CB and four CBD modules in series with $32, 64, 128, 256, 512$ output channels. This block's output features size is $B, 512, 30, 23, 8$, representing batch size $B$, the feature dimensionality, and cylinder height, width and length, respectively.

\subsubsection{3D Max-pooling and Average Pooling}
To aggregate the information, we use max- and average-pooling techniques. The pooling layers compute maximum and average feature values along the spatial/cylindrical dimensions. These layers generate outputs of size $\mathbb R^{B \times 512 }$, which are concatenated to generate the final feature vector of size $\mathbb R^{B \times 1024}$ for further processing, which is discussed in Sec.~\ref{sec:reg}.

\begin{figure}
\begin{center}
\includegraphics[width=0.8\columnwidth]{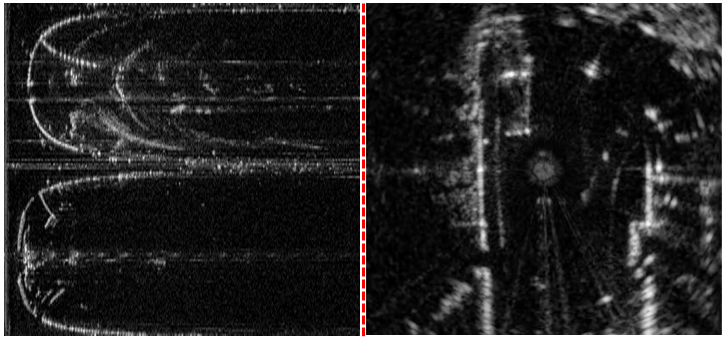}
\caption{Radar sensor output in polar form (\textbf{Left}) and after transformation to Cartesian coordinates (\textbf{Right}).}
\label{fig:radarfig}
\end{center}

\vspace{-6mm}
\end{figure}

\subsection{2D Features Extraction}
\label{sec:2dfeatures}
Here, we explain the 2D feature extraction blocks for image and Radar modalities. The architecture for both blocks in our framework is largely identical, except for an additional 2D convolution layer 
used for the Radar modality, see Fig.~\ref{fig:method}. This additional layer broadcasts channel size from one to three for later processing. We also  transform the polar scanning Radar outputs into Cartesian coordinates as grey-scale images  $\in \mathbb{R}^{ H \times W }$,  as shown in Fig.~\ref{fig:radarfig}. This transformation helps in localization performance. The  architecture for our feature extraction blocks includes a ResNet module, a fine-tuning module and a slot attention-based features filtering module. These components are discussed below in  detail.

\subsubsection{ResNet Blocks}
\label{sec:resnet}
The primary responsibility of this module is to extract useful local features from Radar and camera modalities. The state-of-the-art camera-based localization methods~\cite{wang2020atloc}, \cite{brahmbhatt2018geometry}, \cite{huang2019prior} utilize a pre-trained ResNet model~\cite{he2016deep} as a features extractor. The aforementioned approaches typically involve selecting layers from a significant portion of the pre-trained model, resulting in computationally demanding models. Our framework focuses solely on the ten initial blocks of the pre-trained ResNet-152 model, thereby considerably reducing the computational cost. The input  to this module is  in $\mathbb{R}^{ B \times D \times H \times W }$,  where $B$ is the batch size, $D$ is the input channel size and $(H, W)$ are the height and width of the input. The input values for $D, H, W$ are $3, 512 \text{~and~} 512$, respectively. The output of this module is in $ \mathbb{R}^{B \times 512 \times (H/8) \times (W/8)}$. % with the same dimension order as  input.

%However, these approaches select layers till downstream \SA{What do you mean by downstream} of the model, which makes the localization task computationally expensive. In our framework, we choose the top ten initial blocks of the pre-trained Resnet-152 model
\subsubsection{Fine-tuning Module}
The features extracted in Sec~\ref{sec:resnet} are fine-tuned for localization task in this module. Also, this module makes the features more suitable for the subsequent slot attention-based filtering. In the fine tuning module, the input is passed through a series of 2D convolutional layers, and is augmented with positional information  channel-wise. The resultant features map is flattened along the spatial dimensions and fed into a linear layer for further processing. 
%This module contains a series of sub-modules comprising 2D convolution layers and a max-pooling layer with $stride=2$, a positional encoding layer, a spatial flatten layer and a linear layer. 
The positional encoding in this module is learnable with the encoding tensor of size  $\mathbb{R}^{ B \times (H/32) \times (W/32) \times 1024 }$. The final output size of this module is  $ { B \times (HW/1024) \times 1024}$.

\begin{figure*}
\begin{center}
\includegraphics[width=0.99\textwidth]{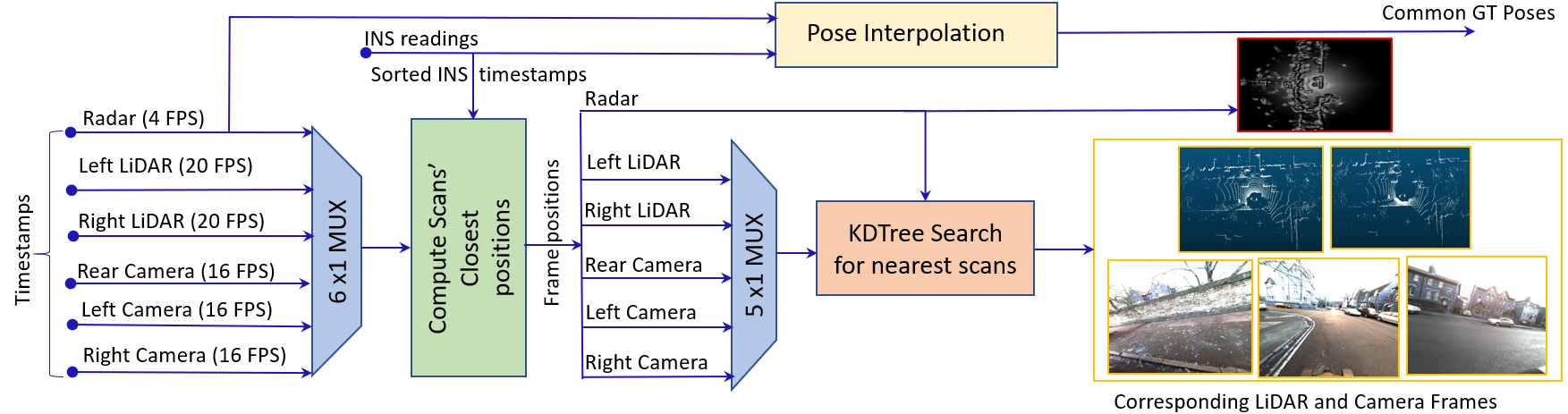}
\vspace{-3mm}
\caption{Schematics for common ground-truth generation for the different types of input sensors namely, Radar, Camera and LiDAR which operate at 4, 16 and 20 frames per second respectively.}
%Common ground truth generation for six different sensors.}

\label{fig:gt}
\end{center}

\vspace{-7mm}
\end{figure*}

\label{sec:slot}
\subsubsection{Slot Attention-based Features Filtering}
Two types of distortions can considerably affect the localization performance. One results from the angular and range errors of the sensors, while the other from the foreground moving objects, e.g., bikes, buses, trucks and pedestrians in the outdoor scene. To minimize these distortions, state-of-the-art methods apply feature filtering techniques. For instance, Barnes~\etal~\cite{barnes2019masking} designed a UNet-type architecture to predict distraction-free Radar odometry. Radarloc~\cite{wang2021radarloc}, AtLoc~\cite{wang2020atloc} and PointLoc~\cite{wang2021pointloc} apply attention-based encoder and decoder techniques to filter out these noises. However, these methods fail to leverage object-centric features in the scene for this purpose, and they are also not  easily tailored to  multiple modalities. To this end, we design a unique slot attention~\cite{locatello2020object} based feature filtering module to  leverage object-centric features with Radar and image modalities. Slot attention is the  key component of our module, whose architectural details are given in  Fig.~\ref{fig:method}. Its primary function is to project the $N$ input features to $K$ output vectors, which represent   slots -  we chose $K = 20$ in our framework. 

%It is an iterative process to create connections between the input features and the slots.
In this module, the slots are randomly initialized, and they  improve iteratively during the training phase. The slots contest for input features through softmax-based attention throughout each iteration, and then use a  gated  recurrent unit (GRU) to update their representation. Similar to a transformer,  \enquote{key}, \enquote{query}, and \enquote{value} vectors are used in the slot attention mechanism. Their computations is denoted by $k(\cdot)$, $q(\cdot)$, and $v(\cdot)$, respectively in the text to follow.  These vectors are kept learnable in our slot attention to map the inputs and slots with a  channel size $D$. 
%However, the traditional transformer relies on multi-headed attention (MHA) modules for a more robust representation; it derives a substantial computational cost, which is aggravated for processing real-world outdoor scenes. We circumvent this bottleneck with the slot-attention mechanism, which is more computationally and memory efficient than MHA. 
The size of the affinity matrix for slot attention is ${ N \times  K }$ as compared to ${ N \times  N }$ for the multi-head attention (MHA) utilized in conventional transformer architectures,  where ${ N >> K }$. This makes slot attention much more efficient then MHA. The slot attention output in $\mathbb{R}^{ B  \times 20 \times 1024 }$ is passed through an MLP and a normal layer for feature refinement. The size of the feature vector generated here is $\in \mathbb{R}^{ B \times 1024}$.

Concretely, a single iteration of the employed slot attention performs the following computation. 
\begin{align}
 \alpha &=\frac{1}{\sqrt{C}} k(input)\cdot q(slots)^{\intercal}  \in \mathbb R^{N \times K}, \\
\Gamma{i,j} &=\frac{e^{\Lambda_{i,j}}}{\sum_{s} e^{\Lambda_{s,j}}  }, \\
\beta &= W^{\intercal}\cdot v(input)  \in \mathbb R^{K \times D},\hspace{2mm}  \\ %\hspace{2mm} 
W &= \frac{ \Gamma{i,j}} {\sum_{s=1}^{N}{\Gamma{s,j}}}.
\end{align}
In the above, $\alpha$, $\Gamma$, and $\beta$ respectively represent the attention coefficient matrix, normalized attention over the slots, and the slot update for further processing by the GRU. Finally, a GRU with as many hidden units as the dimensionality of the slots is used to update the slot. The update is based on a previous slot state and the signal $\beta$.

\label{sec:modencoding}
\subsection{Modality Encoding}
Inspired by the transformer’s positional encoding technique, we propose modality encoding and incorporate it in our network to identify the sensor modality. Unlike  pre-fixed `cosine’ and `sine’ positional encoding, we learn the modality encoding for optimal performance. Since our framework supports three modalities, i.e., image, Radar and point cloud, we randomly initialize three modality encoding vectors with the same size as the feature  vectors. These vectors are learned during the training phase and are added to the feature vectors to detect the sensor modalities. Experiments showed that this modality encoding works well during inference time.

\label{sec:reg}
\subsection{Regression Module}
The feature vectors encoded in the earlier stages of the model are  passed to a Regression Module which is responsible to predict the 6DoF pose for localization. It comprises two common fully connected (FC) layers at the top, preceded by two parts of four FC layers - see Fig.~\ref{fig:method}. The network has a dicephalous architecture to precisely predict the translation and rotation parameters for the 6DoF pose.  The channel output sizes in each division of the FC layers are $1024, 512, 256, 3$. The initialization of the layers is set with  Xavier\_uniform distribution and ReLU activation is used. To reduce the variations between rotation and translation values, the rotation branch of this module is normalized.

\label{sec:loss}
\subsection{Training Loss Computation}
We  utilize  $\ell_1$-loss for rotation and translation, which we found more suitable for our network due to the various types of input data. We combine the rotation and translation losses to compute the loss for each input data with learnable balancing factors $\alpha$  and $\beta$, as shown in Eq.~(\ref{eq3}). During the training phase, we forward pass inputs from all six sensors and combine the loss for each input to compute the net loss for the network, see Eq.~(\ref{eq4}). Finally, the net loss is back propagated in the network for optimization.
\begin{flalign}
\label{eq3}
\mathcal L &=   \| t - t'\| e^{\alpha} +  \alpha +   \| r - r'\|e^{\beta} +\beta.
\end{flalign}
\begin{flalign}
\label{eq4}
\mathcal L_{net} &=  \mathcal L_{L1} +  \mathcal L_{L2} +  \mathcal L_{C1} +  \mathcal L_{C2} +  \mathcal L_{C3} +  \mathcal L_{R}.
\end{flalign}
In Eq.~(\ref{eq3}), $t$ and $t'$ indicate ground-truth and predicted translation, whereas $r$ and $r'$ denote the respective rotations.  
%$(t',r')$, $(t,r)$ are individual sensor loss, the ground-truth and predicted poses, i.e., translations $t$ (along X, Y, Z directions) and rotations $r$ (in yaw, roll, pitch). 
In Eq.~(\ref{eq4}), $\mathcal L_{net} $,  $ \mathcal  L_{L1}$, $\mathcal L_{L2}$, $ \mathcal L_{C1}$, $\mathcal L_{C2}$, $\mathcal L_{C3}$, $\mathcal  L_{R}$ are the net loss, LiDAR left, LiDAR right, camera left, camera right, camera rear, and Radar sensor losses, respectively.

\begin{table*}[!t]
\caption{%Results on the Oxford Radar RobotCar. The values represent the mean rotation (in degrees) and translation (in meter) errors. Our method achieves the lowest average errors for all test sequences.
Mean translation (meters) and rotation (deg.) errors on the Radar RobotCar dataset\cite{RadarRobotCarDatasetICRA2020} compared to Radar SLAM\cite{hong2020radarslam}, adapted AtLoc\cite{wang2020atloc}, RadarLoc\cite{wang2021radarloc}, AtLoc\cite{wang2020atloc}, MapNet\cite{brahmbhatt2018geometry}, PoseNet\cite{kendall2015posenet}, DCP\cite{wang2019deep}, VLAD\cite{uy2018pointnetvlad} and PointLoc\cite{wang2021pointloc} . %Our method performs the best.
} 
\vspace{-3mm}

%translation error on all routes and lowest yaw error on 3 routes.}
\renewcommand{\arraystretch}{1.1} % Default value: 1

\renewcommand{\tabcolsep}{2.5pt}
\centering
\resizebox{1\textwidth}{!}{
\begin{tabular}{c|c c c c|c c c c| c c  c c }
%\thickhline
 \hline
  \multicolumn{5}{c|}{Radar}& \multicolumn{4}{c|}{Camera} &\multicolumn{4}{c}{LiDAR} \\ \cline{1-13}
 
\multirow{3}{*} {Sq} & Radar & Adapted &\multirow{2}{*}{RadarLoc} & {UnLoc}& \multirow{2}{*}{AtLoc} & \multirow{2}{*}{MapNet} & \multirow{2}{*}{PoseNet }& {UnLoc} & \multirow{2}{*}{DCP}  & {PointNet} & \multirow{2}{*}{PointLoc} &{UnLoc}\\
&SLAM & AtLoc & & Ours& & & &Ours & & VLAD& &Ours \\  \hline

%& Radar &Radar  & Radar & Camera & Camera&Camera & LiDAR &LiDAR&LiDAR & Multi-sensors\\ 

 % &a&b&  & & c & d &  f&  g& h & j&&}\\
 6 & 49.8, 5.2°  & 15.9, 4.2° &  8.4, 3.4° &  \textbf{2.92}, \textbf{1.26°} & 15.4, 3.4°  & 32.2, 5.4°  & 51.1, 6.4°& \textbf{2.91},  \textbf{1.27°} & 18.5, 2.1°  & 28.9, 5.2°&14.4, 2.8° &\textbf{1.71}, \textbf{0.68°}\\

7 & 24.7, 3.4°& 13.2, 3.8° & 5.1, 2.9° & \textbf{2.82}, \textbf{1.68°} & 39.7, 8.3° & 47.8, 5.4°& 80.3, 6.5°  & \textbf{2.83}, \textbf{1.69°}& 02.8, 1.7° & 17.6, 3.9° &08.5, 1.8° &\textbf{1.67}, \textbf{0.68°}\\

8 &26.1, 1.6° &14.2, 2.9° &6.6, 3.1° &\textbf{2.85}, \textbf{1.21°}&  31.7, 4.3° &51.9, 7.7°  &111, 12.8°  & \textbf{2.85}, \textbf{1.20°}& 16.4, 2.3°&23.6, 5.9° &09.5, 2.1° &\textbf{1.57}, \textbf{0.61°}\\

9 &39.8, 5.7° &15.7, 3.2°  &6.5, 2.9° &\textbf{2.84}, \textbf{1.49°} & 47.1, 9.4°&14.9, 2.8°  &45.5, 4.0° & \textbf{2.83}, \textbf{1.50°}& 13.6, 1.9°&13.7, 2.6°&11.5, 2.0° &\textbf{1.51}, \textbf{0.35°} \\

10 &17.8, \textbf{1.7°} &13.2, 1.9° & 5.3, 1.8°& \textbf{2.67}, 2.04°&10.4, \textbf{1.3°} & - &- & \textbf{2.68}, 2.05° & - &- & 08.4, 1.4° &\textbf{1.57},  \textbf{0.51°}\\
\hline 
Av & 31.7, 3.5° & 14.4, 3.2°  & 6.4, 2.8°&  \textbf{2.82},  \textbf{1.53°} & 28.8, 5.3° &36.7, 5.4° & 72.0, 7.4°&  \textbf{2.82},  \textbf{1.54°} &15.8, 2.1° &20.8, 4.4° & 10.5, 2.0°& \textbf{1.61}, \textbf{0.57°}\\

\hline
\end{tabular}}
\label{tab:table1}
% \vspace{-5mm}
\end{table*}

\begin{table*}[!t]
\caption{Ablation study on test sequences of Oxford Radar RobotCar. Mean translation (meters) and rotation (deg.) errors are reported for the 3 modalities separately and in combinations. L1/L2, C1/C2/C3 \& R stand for LiDAR left/right, camera left/right/rear \& Radar. We report results for C1 only since all three cameras give approximately similar accuracy.} %translation error on all routes and lowest yaw error on 3 routes.}

\centering
\resizebox{16cm}{!}{
\begin{tabular}{c| c c c c c c  c  c }
%\thickhline
 \hline

{Seq} & L1 & L2 & C1 & R   &  L1, C1, R  & L2, C2, R &   L1, L2, R &  All \\

% \multirow{2}{*} {Sensor} & LiDAR & LiDAR & Cameras &{Mix1} & Mix2 & \multirow{2}{*}{DCP\cite{wang2019deep}}  & {PointNet} & \multirow{2}{*}{PointLoc\cite{wang2021pointloc} }& \multirow{2}{*}{Ours}\\
% &\scriptsize{left} & \scriptsize{left\& right}& \scriptsize{left, right \& rear} & \scriptsize{$LiDAR_{left}$, $Cam_{left}$ \& Radar} & & & & VLAD\cite{uy2018pointnetvlad}& &\\ 

\hline
 % &a&b&  & & c & d &  f&  g& h & j&&}\\
 6 & 1.75, 0.77°  & 1.71, 0.68° & 2.91, 1.27° & 2.92, 1.26°  & 1.68, 0.64°  & 1.62, 0.59°& 1.50, 0.58°  & {1.49}, {0.57°}\\
7 & 1.69, 0.61°  & 1.67, 0.68° & 2.83, 1.69° & 2.82, 1.68°  & 1.63, 0.52°  & 1.60,0.61°& 1.47, 0.50°  & {1.46}, {0.49°}\\

8 & 1.58, 0.65° & 1.57, 0.61° & 2.85, 1.20° & 2.85, 1.21° & 1.52, 0.59°  & 1.51, 0.57°& 1.38, 0.55°  & 1.37, 0.54°\\

9 & 1.55, 0.37°  &1.51, 0.35° & 2.83, 1.50° & 2.84, 1.49°  & 1.49, 0.36°  & 1.47, 0.35°& 1.34, 0.33°  & {1.33}, {0.32°}\\
10 &1.52, 0.56°  & 1.57, 0.51° & 2.68, 2.05° & 2.67, 2.04°  & 1.47, 0.56°  & 1.52, 0.48°& 1.38, 0.50°  & 1.36, 0.48°\\
\hline
Avg & 1.62, 0.59° & 1.61, 0.57°  & 2.82, 1.54°&2.82, 1.53° &1.56, 0.53° & 1.54, 0.52°& 1.42, 0.49° &   \textbf{1.40}, \textbf{0.48°}\\
\hline
\end{tabular}}
\label{tab:table2}
\vspace{-5mm}
\end{table*}

\section{EXPERIMENTS}
\vspace{-1mm}
We evaluate the proposed method for the localization task on benchmark datasets and compare it with state-of-the-art techniques. To ensure a fair comparison, we choose popular existing methods based on the availability of source code by the original authors. We present results on three major localization datasets: Oxford Radar RobotCar~\cite{RadarRobotCarDatasetICRA2020},  Apollo-SouthBay~\cite{L3NET_2019_CVPR} and Perth-WA~\cite{ibrahim2023slice}. Our results establish the effectiveness of the proposed model for point cloud, image and Radar modality, both individually and collectively. Prior to presenting the results for each dataset, we discuss implementation details in the section below.
 \vspace{-2mm}

\subsection{Implementation Details}
\vspace{-1mm}
To ensure fair benchmarking, we adopt uniform configurations for our method across the Oxford Radar RobotCar, Apollo-SouthBay and Perth-WA datasets. We apply batch sizes of 6 and 1 for training and testing, respectively. We use Adam optimizer with a learning rate of $0.0001$ and weight decay $0.0005$. At the outset, the model is trained on Oxford Radar RobotCar for multi-modalities and then fine-tuned on the Apollo-SouthBay dataset and Perth-WA dataset for point cloud modality. The model is trained for 40 epochs on all three datasets. For all experiments, NVIDIA GeForce RTX 3090 GPU with 24 GB memory is used. The experiments are conducted using PyTorch 1.8.0 on Ubuntu 18.04 OS.

%The post-processing is done to pinpoint the query frame in the 3D point cloud map after 6DoF pose prediction. A rotation matrix is computed based on the predicted pose values. While the query frame is first down-sampled before aligning it with the map. Lastly, the processed query frame is multiplied with the rotation matrix to visualize the location of the frame in the map.

\subsection{Results on Oxford Radar RobotCar}
%\subsubsection{Dataset details}
%\vspace{2mm}
\noindent
\textbf{Dataset details:}
The Oxford Radar RobotCar%\st{dataset}
~\cite{RadarRobotCarDatasetICRA2020} is an extension %\st{of The Oxford RobotCar dataset} 
of their previous dataset~\cite{RobotCarDatasetIJRR} that includes three different modalities: RGB camera images, Radar data, and point clouds from six sensors: left, right, and rear cameras; a Navtech CTS350-XFMCW Radar scanner; and left and right Velodyne HDL-32E LiDAR. NovAtel SPAN-CPT ALIGN inertial (INS) and GPS navigation systems are used to collect the ground-truth poses for this dataset. It covers a total of 280km of urban area, including more than 30 sequences, each captured over 9km. This dataset is large and challenging for localization due to the presence of a variety of foreground objects, such as people and cars. We use the same training and test sequences as Radarloc~\cite{wang2021radarloc} for our experiments on this dataset.
%\textcolor{red}{[XX]} \SA{isn't both of them the same datasets?}They are different.

\vspace{2mm}
\noindent
\textbf{Common Ground-truth Generation for multi-sensors:}
%\subsubsection{Common Ground-truth Generation for multi-sensors}
Our approach has the unique ability to leverage all three data modalities provided by the dataset. 
However, the frame rates for Radar (4Hz), LiDAR  (20Hz), and camera (16Hz) data have a large disparity between them, which leads to timestamp misalignment between the modality sensors and the ground-truth. 
%\st{Resolving this issue is a contribution of this paper.} 
To generate a unified ground-truth for all the data sensors at a given time, we synchronize the Radar timestamps with the ground-truth poses using interpolation between GPS/INS measurements and the Radar timestamps. This step of ours provides %\st{gives us} 
%\SA{of ours provide or of ours furnish} 
ground-truth 
%\SA{be consistent: either use \enquote{ground truth} or \enquote{ground-truth}} 
pose for each Radar frame. We then compute the position information for each frame for all modality sensors based on their timestamps. To acquire the corresponding frame for each remaining sensor, we search for the closest frame position corresponding to each Radar frame using the minimum Euclidean distance with KDTree search~\cite{scikit-learn}. Each Radar frame and its corresponding closest searched frame share the same ground-truth. Missing GPS/INS data is handled by interpolating values from visual odometry data provided by the Radar RobotCar. In this way, we calculate single ground-truth pose for each frame of all modality sensors. The process is also illustrated in Fig.~\ref{fig:gt}.

\vspace{2mm}
\noindent
\textbf{Performance:}
%\subsubsection{Results}
%\st{Our} \SA{We present the} 
We present the experimental results %\st{are presented} 
in Table~\ref{tab:table1} where we use individual data modalities to compare with the approaches of the respective modalities. Due to the universal nature of our method, we are able to compare %\st{our method} 
with  Radar, camera, and LiDAR-based 
%\SA{be consistent, either start all with capital or use small} 
deep localization methods. Our technique outperforms all the existing methods by a considerable margin, which is clear from the respective sections of the table. For the point cloud modality, our approach surpasses the best performer PointLoc~\cite{wang2021pointloc} by reducing the errors for translation and rotation nearly by \emph{$5\times$} and \emph{$3\times$}, respectively.
%\st{five and three times} \SA{by \emph{$5\times$} and \emph{$3\times$}, respectively}. %reducing them by around five and three times respectively, which demonstrates the efficacy of our proposed method.
The results confirm that our encoding and sparse 3D convolutional modules %\st{module} \SA{modules} 
are effective components for  localization with point clouds. Table~\ref{tab:table1} also ascertains that our  method is much more accurate than the camera and Radar-based methods. Our technique outperforms RadarLoc~\cite{wang2021radarloc} by a more than $4\times$ %\st{4 times} \SA{\emph{$4\times$}} 
error reduction in translation and rotation estimates. We conjecture that the strong performance of our %\st{technique} \SA{approach} 
approach has two main sources. Firstly, for each modality, our network is carefully designed with the state-of-the-art representation learning components. 
%\st{the} stat\SA{e}-of-the-art representation learning components \st{for the modality}. 
Secondly, our model is able to leverage the complementary information from different modalities during the learning stage to better train each individual modality network. For each modality, the inference stage is able to %\st{is able to} \SA{can}
take guidance from learned positional encoding for optimal performance.

 \begin{table*}[h]
\caption{Results on  Apollo-SouthBay dataset. The values represent RMSE for rotation (deg) and translation (cm). Our method achieves the lowest average errors for three routes.}
\vspace{-2mm}

\centering
\renewcommand{\tabcolsep}{8pt}
\resizebox{16cm}{!}{
\begin{tabular}{l |l c  c| c|| c c c |c c c}
% \thickhline
 \hline
\multicolumn{1}{c|}{Methods}    & $Yaw$ &  $Roll$   & $Pitch$   & $Rot$  & $X$  & $Y$   & $Z$  & $Trans$  &\\ \hline
Levinson~\etal~\cite{levinson2010robust} & - & -  &-  & - & $11.9$ &  $9.3$ & $4.6$   & $8.9$ & \\

  Wan~\etal~\cite{wan2018robust} & $3.8^{\circ}$ &-  & - & - & $ 5.0$ & $ 3.6$ & $\mathbf{2.6}$  & $3.7$ & \\ 
  
  L3-net ~\cite{lu2019l3} &  $\mathbf{1.6^{\circ}}$ & -  & - &-  & ${5.0}$ & ${3.6}$ & ${2.7}$  & ${3.8}$& \\ 

 Slice3D ~\cite{ibrahim2023slice} &  ${3.2^{\circ}}$ &  ${6.7^{\circ}}$ & ${7.3^{\circ}}$ & ${6.4^{\circ}}$ & $\mathbf{2.4}$ & ${3.3}$ & ${3.1}$  & ${2.9}$& \\ 

  UnLoc (Ours)  & ${2.8}^{\circ}$ & $\mathbf{1.3^{\circ}}$ & $\mathbf{1.4^{\circ}}$ & $\mathbf{1.83}^{\circ}$ & ${3.1}$ & $\mathbf{2.7}$ &     $\mathbf{2.6}$ & $\mathbf{2.8}$& \\ \hline

\end{tabular}}
\label{tab:table4}

\end{table*}

 \begin{table*}[h]
\caption{Results on Perth-WA dataset. The values represent the absolute mean error for rotation (degrees) and translation (meters). Our method has the least error in all cases.}
\vspace{-2mm}

\centering
\renewcommand{\tabcolsep}{6pt}
\resizebox{16cm}{!}{
\begin{tabular}{l |l c  c| c|| c c c |c c c}
% \thickhline
 \hline
\multicolumn{1}{c|}{Methods}    & $Yaw$ &  $Roll$   & $Pitch$   & $Rot$  & $X$  & $Y$   & $Z$  & $Trans$  &\\ \hline
 PointLoc~\cite{wang2021pointloc} & $0.26^{\circ}$ &  $1.96^{\circ}$ & $0.15^{\circ}$ & $0.75^{\circ}$ & $29.70$ &  $37.49$ & $7.80$   & $25.00$ & \\

  Slice3D (baseline)~\cite{ibrahim2023slice} & $0.32^{\circ}$ & $2.42^{\circ}$ & $0.27^{\circ}$ & $1.00^{\circ}$ & $ 14.20$ & $ 17.05$ & $8.50$  & $13.25$ & \\ 

 Slice3D (pre-trained)~\cite{ibrahim2023slice} &  ${0.17^{\circ}}$ &  ${1.52^{\circ}}$ & $\mathbf{0.10^{\circ}}$ & ${0.59^{\circ}}$ & ${6.26}$ & ${06.55}$ & ${2.86}$  & ${5.23}$& \\ 

  UnLoc (Ours)  & $\mathbf{0.12}^{\circ}$ & $\mathbf{1.13^{\circ}}$ & $0.23^{\circ}$ & $\mathbf{0.49}^{\circ}$ & $\mathbf{2.11}$ & $\mathbf{01.91}$ &     $\mathbf{1.13}$ & $\mathbf{1.72}$& \\ \hline

% RangeNet++~\cite{milioto2019rangenet++} & 65.5 & 66.0 & 21.3 & 77.2 &  80.9 & 30.2&  66.8 & 69.6 & 52.1 & 54.2 & 72.3 &  94.1 & 66.6 &  63.5 &  70.1 &  83.1 & 79.8\\

% PolarNet~\cite{zhang2020polarnet} &  71.0 & 74.7 & 28.2 & 85.3 & 90.9 & 35.1 & 77.5 & 71.3 & 58.8 & 57.4 & 76.1 & \underline{96.5} & 71.1 & 74.7 & 74.0 & 87.3 & 85.7\\

% Salsanext~\cite{cortinhal2020salsanext} &72.2& \underline{74.8} &34.1& 85.9& 88.4& 42.2 &72.4 &72.2& 63.1& 61.3 &76.5& 96.0 &70.8& 71.2& 71.5& 86.7 &84.4\\

% Cylinder3D~\cite{zhu2020cylindrical} & \underline{76.1} & \textbf{76.4} & \underline{40.3} & \underline{91.2}& \textbf{93.8} &\underline{51.3}& \textbf{78.0}& \underline{78.9} & \textbf{64.9}& \underline{62.1}& \underline{84.4} &\textbf{96.8} & \underline{71.6} & \textbf{76.4} & \underline{75.4} & \textbf{90.5}& \underline{87.4}\\\hline \hline 
% SAT3D (Ours) & \textbf{76.7} & 74.3   &  \textbf{43.9}   &    \textbf{91.3}   &    \underline{92.8}   &    \textbf{53.0}   &    \underline{77.8}   &    \textbf{79.7}   &   \underline{64.8}   &    \textbf{63.3}   &    \textbf{85.7}   &    \underline{96.5}   &    \textbf{73.4}   &    \underline{74.7}   &   \textbf{77.3}   &    \underline{89.9}   &    \textbf{88.0}   \\  
% \hline
% \thickhline
\end{tabular}}
\label{tab:table3}
\vspace{-4mm}

\end{table*}

\noindent
\textbf{Ablation studies:}
%\subsubsection{Ablation studies}
To investigate the influence of different blocks of our technique, we conduct ablation studies and summarize the results in Table~\ref{tab:table2}. For the experiments, we kept the architecture for each block the same but turned on/off the image, Radar, and point cloud modality blocks to determine the impact on the localization results. First, we test our framework with a single modality and turned off the other two modality blocks. Table~\ref{tab:table2} reports the full results of these experiments in the first four columns, which can be compared with the results in Table~\ref{tab:table1}.
%that our proposed method for single modality surpasses existing state-of-the-art localization methods for image, radar, and point cloud modalities with a significant margin in terms of the localization error. 
Among the single modalities, our approach already  achieves the best overall performance. 
%highest localization accuracy for the point cloud modality due to its rich 3D  information. 
%The results for Radar and image modalities are still much better than existing localization techniques.
%We achieved the best performance for the left LiDAR sensor with 1.61 and 0.57 translation and rotation errors, respectively. 
To further analyse the performance of our network, we  test different combinations of the blocks in our method using three sensors at a time. % to find the best network performance. 
From Table~\ref{tab:table2}, it is clear that the localization performance of our technique improves by using different data modalities. 
%The localization errors for translation and rotation decreased in these trials. 
In this case, the best performance is achieved with a combination of left Lidar, right Lidar and Radar, resulting in 1.42m and 0.49$^\circ$ errors for translation and rotation, respectively.
Finally, we utilized all six data streams in our technique. The last column of the table shows that this results in the overall best performance for our technique. This ascertains that each data modality is able to contribute to improve the performance, and that our network is able to leverage the complementary data information effectively. 
%The best localization accuracy was obtained when we passed the data from all six sensors to the network, as seen in Table~\ref{tab:table2}. 
%This demonstrates that our framework offers the best localization performance for all modalities due to its multi-sensor data training and its use of 3D sparse convolution and slot attention-based feature filtering modules.

% \makeatletter
% \def\thickhline{%
%   \noalign{\ifnum0=`}\fi\hrule \@height \thickarrayrulewidth \futurelet
%   \reserved@a\@xthickhline}
% \def\@xthickhline{\ifx\reserved@a\thickhline
%               \vskip\doublerulesep
%               \vskip-\thickarrayrulewidth
%              \fi
%       \ifnum0=`{\fi}}
% \makeatother

%%%%%%%%%%%%%%%%%%%%%%%%%%%%%%%%

% \begin{figure}
% \includegraphics[width=1\columnwidth]{TranslationAndRotationError.PNG}
% \caption{\bf{Left} image shows the mean translation error and \bf{right} image represents the mean rotation error of our proposed localization method on test frames of our proposed localization dataset.}
% \label{fig:error}
% \end{figure}

\subsection{Results on the Apollo-SouthBay Dataset}
%\subsubsection{Dataset}
%\vspace{2mm}
\noindent
\textbf{Dataset:}
%The ApolloSouthBay~\cite{L3NET_2019_CVPR} is a comprehensive localization dataset collected in San Francisco, USA. An IMU based system is utilized to record the ground-truth poses for the LiDAR frames. The dataset includes six routes: BaylandsToSeafood, ColumbiaPark, Highway237, MathildaAVE, SanJoseDowntown, and SunnyvaleBigLoop. The dataset offers training and test sets for each route. The dataset is captured in residential, urban, downtown area and highways. All the routes are used for the training set, while four routes are employed for the test set.
ApolloSouthBay~\cite{L3NET_2019_CVPR} is a comprehensive localization dataset collected in San Francisco, USA, utilizing an IMU-based system to record the ground-truth poses for the LiDAR frames. The dataset is captured in residential, urban, downtown area and highways. The dataset includes six routes: BaylandsToSeafood, ColumbiaPark,  SanJoseDowntown, SunnyvaleBigLoop, Highway237 and MathildaAVE. All these routes provide separate training and test sets. 
%The dataset is captured in residential, urban, downtown area and highways, and it offers training and test sets for each route. All the routes are used for the training set, while four routes are employed for the test set.
%The dataset offers training and test sets for each route and captures the highways, residential, urban, and downtown areas where the training set comprises all the routes, while the test set consists only of four. %\SA{I have re-wrote this section, please check}

\vspace{2mm}
\noindent
\textbf{Performance:}
%\subsubsection{Results}
The outcomes of our experiments on this dataset are presented in Table~\ref{tab:table4}. For benchmarking, we fine-tune our model on the training sets and assess our model on all six routes of the test set. 
We employ RMSE as the evaluating metric by following~\cite{L3NET_2019_CVPR} and  
%as shown in Table~\ref{tab:table4}. 
compare our approach with the state-of-the-art localization methods, Levinson~\etal~\cite{levinson2010robust}, Wan~\etal~\cite{wan2018robust}, L3-net \cite{L3NET_2019_CVPR} and Slice3D \cite{ibrahim2023slice}. In the Table, we report average values on all six routes. Levinson~\etal, Slice3D~\cite{ibrahim2023slice} and L3-net~\cite{L3NET_2019_CVPR}are single modality methods, whereas Wan~\etal \cite{wan2018robust} is a fusion model that integrates multiple sensors including a GPS. The results of compared methods are  taken directly from the literature. %\SA{it is taken from literature or their respective papers}. 
Our method outperforms all techniques by achieving the lowest average errors across the six routes. We avoid reporting results of individual routes for brevity, however, note that our method achieves the best performance on each individual route as well. %Our results indicate that our method with multi-modality capabilities provides more robust localization in the outdoor environment.

\subsection{Results on Perth-WA dataset}
%\subsubsection{Dataset}
%\vspace{2mm}
\noindent
\textbf{Dataset:}
Perth-WA dataset~\cite{ibrahim2023slice} is captured in the Central Business District (CBD), Perth, Western Australia. The dataset comprises a LiDAR map of 4km$^2$ with 6DoF ground-truth per frame. The scenes include commercial structures, residential areas, food streets, complex routes, hospital buildings \etc The data is  collected in three different two-hour sessions under various weather conditions. We apply the same split for training and testing sets as in~\cite{ibrahim2023slice}. The training set comprises 20K frames of sparse and dense point clouds, and another 2.2K frames are used as the test set. \textcolor{black}{The dataset is available online on IEEE data portal~\cite{s2p2-2e66-23}}
%\SA{it is unclear to me what you mean. You chose 20k for training, and then you chose 2200 for testing. where the testing frames come from? }, taken from different regions of Perth CBD.

%\subsubsection{Results}
\vspace{2mm}
\noindent
\textbf{Performance:}
We evaluate the performance of our approach against recent point cloud-based localization approaches: PointLoc~\cite{wang2021pointloc}, Slice3D baseline and pretrained models of~\cite{ibrahim2023slice}, as shown in Table~\ref{tab:table3}.
%\SA{which table? do you mean Table~\ref{tab:table3}? }. 
To conduct this experiment, we fine-tune our framework on the training set and evaluate it on the test set. In line with Pointloc~\cite{wang2021pointloc}, we use the Mean Absolute Error of poses for analyzing the performance. 
%Table~\ref{tab:table3} summarizes the results for this experiment \SA{remove this sentence if you have mentioned Table~\ref{tab:table3} before}. 
Our localization approach outperforms all the methods for angular and translational mean error values. These results show that our proposed approach facilitates more effective point cloud feature learning, making it a preferred choice for outdoor localization using LiDAR frames.

\section{CONCLUSION}
\vspace{-1mm}
This paper presents a novel localization framework for multi-sensors along with a deep neural network architecture that processes LiDAR, Radar and/or camera inputs on demand. 
The proposed network employs 3D sparse convolution and cylindrical partition to process LiDAR frames, and implements ResNet blocks with fine-tuning layers and a slot attention-based feature filtering module for the Radar and image modalities. It also introduces a novel learnable modality encoding technique to identify the type of input data modality. The network is trained on six inputs from three sensor types and can process either a single or multiple sensor inputs at inference. This makes our method robust to sensor failure. Our method is useful for self-driving vehicles that need precise localization regardless of the weather conditions. 
We benchmark our method on three benchmark datasets and achieve state of the art results. 

\bibliographystyle{./IEEEtran} % use IEEEtran.bst style
\bibliography{./mybib}

\begin{thebibliography}{10}
\providecommand{\url}[1]{#1}
\csname url@rmstyle\endcsname
\providecommand{\newblock}{\relax}
\providecommand{\bibinfo}[2]{#2}
\providecommand\BIBentrySTDinterwordspacing{\spaceskip=0pt\relax}
\providecommand\BIBentryALTinterwordstretchfactor{4}
\providecommand\BIBentryALTinterwordspacing{\spaceskip=\fontdimen2\font plus
\BIBentryALTinterwordstretchfactor\fontdimen3\font minus
  \fontdimen4\font\relax}
\providecommand\BIBforeignlanguage[2]{{%
\expandafter\ifx\csname l@#1\endcsname\relax
\typeout{** WARNING: IEEEtran.bst: No hyphenation pattern has been}%
\typeout{** loaded for the language `#1'. Using the pattern for}%
\typeout{** the default language instead.}%
\else
\language=\csname l@#1\endcsname
\fi
#2}}

\bibitem{segal2009generalized}
A.~Segal, D.~Haehnel, and S.~Thrun, ``Generalized-icp.'' in \emph{Robotics:
  science and systems}, vol.~2, no.~4.\hskip 1em plus 0.5em minus 0.4em\relax
  Seattle, WA, 2009, p. 435.

\bibitem{stoyanov2012fast}
T.~Stoyanov, M.~Magnusson, H.~Andreasson, and A.~J. Lilienthal, ``Fast and
  accurate scan registration through minimization of the distance between
  compact 3d ndt representations,'' \emph{The International Journal of Robotics
  Research}, vol.~31, no.~12, pp. 1377--1393, 2012.

\bibitem{ibrahim2023slice}
M.~Ibrahim, N.~Akhtar, S.~Anwar, M.~Wise, and A.~Mian, ``Slice transformer and
  self-supervised learning for 6dof localization in 3d point cloud maps,''
  \emph{arXiv preprint arXiv:2301.08957}, 2023.

\bibitem{wang2021pointloc}
W.~Wang, B.~Wang, P.~Zhao, C.~Chen, R.~Clark, B.~Yang, A.~Markham, and
  N.~Trigoni, ``Pointloc: Deep pose regressor for lidar point cloud
  localization,'' \emph{IEEE Sensors Journal}, vol.~22, no.~1, pp. 959--968,
  2021.

\bibitem{kendall2015posenet}
A.~Kendall, M.~Grimes, and R.~Cipolla, ``Posenet: A convolutional network for
  real-time 6-dof camera relocalization,'' in \emph{Proceedings of the IEEE
  international conference on computer vision}, 2015, pp. 2938--2946.

\bibitem{kendall2016modelling}
A.~Kendall and R.~Cipolla, ``Modelling uncertainty in deep learning for camera
  relocalization,'' in \emph{2016 IEEE international conference on Robotics and
  Automation (ICRA)}.\hskip 1em plus 0.5em minus 0.4em\relax IEEE, 2016, pp.
  4762--4769.

\bibitem{cen2018precise}
S.~H. Cen and P.~Newman, ``Precise ego-motion estimation with millimeter-wave
  radar under diverse and challenging conditions,'' in \emph{2018 IEEE
  International Conference on Robotics and Automation (ICRA)}.\hskip 1em plus
  0.5em minus 0.4em\relax IEEE, 2018, pp. 6045--6052.

\bibitem{wang2021radarloc}
W.~Wang, P.~P. de~Gusmao, B.~Yang, A.~Markham, and N.~Trigoni, ``Radarloc:
  Learning to relocalize in fmcw radar,'' in \emph{2021 IEEE International
  Conference on Robotics and Automation (ICRA)}.\hskip 1em plus 0.5em minus
  0.4em\relax IEEE, 2021, pp. 5809--5815.

\bibitem{RadarRobotCarDatasetICRA2020}
\BIBentryALTinterwordspacing
D.~Barnes, M.~Gadd, P.~Murcutt, P.~Newman, and I.~Posner, ``The oxford radar
  robotcar dataset: A radar extension to the oxford robotcar dataset,'' in
  \emph{Proceedings of the IEEE International Conference on Robotics and
  Automation (ICRA)}, Paris, 2020. [Online]. Available:
  \url{https://arxiv.org/abs/1909.01300}
\BIBentrySTDinterwordspacing

\bibitem{L3NET_2019_CVPR}
W.~Lu, Y.~Zhou, G.~Wan, S.~Hou, and S.~Song, ``L3-net: Towards learning based
  lidar localization for autonomous driving,'' in \emph{Proceedings of the IEEE
  Conference on Computer Vision and Pattern Recognition}, 2019, pp. 6389--6398.

\bibitem{elhousni2020survey}
M.~Elhousni and X.~Huang, ``A survey on 3d lidar localization for autonomous
  vehicles,'' in \emph{2020 IEEE Intelligent Vehicles Symposium (IV)}.\hskip
  1em plus 0.5em minus 0.4em\relax IEEE, 2020, pp. 1879--1884.

\bibitem{gao2021fully}
X.~Gao, Q.~Wang, H.~Gu, F.~Zhang, G.~Peng, Y.~Si, and X.~Li, ``Fully automatic
  large-scale point cloud mapping for low-speed self-driving vehicles in
  unstructured environments,'' in \emph{2021 IEEE Intelligent Vehicles
  Symposium (IV)}.\hskip 1em plus 0.5em minus 0.4em\relax IEEE, 2021, pp.
  881--888.

\bibitem{kovalenko2019sensor}
D.~Kovalenko, M.~Korobkin, and A.~Minin, ``Sensor aware lidar odometry,'' in
  \emph{2019 European Conference on Mobile Robots (ECMR)}.\hskip 1em plus 0.5em
  minus 0.4em\relax IEEE, 2019, pp. 1--6.

\bibitem{nubert2021self}
J.~Nubert, S.~Khattak, and M.~Hutter, ``Self-supervised learning of lidar
  odometry for robotic applications,'' in \emph{2021 IEEE International
  Conference on Robotics and Automation (ICRA)}.\hskip 1em plus 0.5em minus
  0.4em\relax IEEE, 2021, pp. 9601--9607.

\bibitem{dube2018segmap}
R.~Dub{\'e}, A.~Cramariuc, D.~Dugas, J.~Nieto, R.~Siegwart, and C.~Cadena,
  ``Segmap: 3d segment mapping using data-driven descriptors,'' \emph{arXiv
  preprint arXiv:1804.09557}, 2018.

\bibitem{vaswani2017attention}
A.~Vaswani, N.~Shazeer, N.~Parmar, J.~Uszkoreit, L.~Jones, A.~N. Gomez,
  L.~Kaiser, and I.~Polosukhin, ``Attention is all you need,'' \emph{arXiv
  preprint arXiv:1706.03762}, 2017.

\bibitem{brahmbhatt2018geometry}
S.~Brahmbhatt, J.~Gu, K.~Kim, J.~Hays, and J.~Kautz, ``Geometry-aware learning
  of maps for camera localization,'' in \emph{Proceedings of the IEEE
  conference on computer vision and pattern recognition}, 2018, pp. 2616--2625.

\bibitem{walch2017image}
F.~Walch, C.~Hazirbas, L.~Leal-Taixe, T.~Sattler, S.~Hilsenbeck, and
  D.~Cremers, ``Image-based localization using lstms for structured feature
  correlation,'' in \emph{Proceedings of the IEEE International Conference on
  Computer Vision}, 2017, pp. 627--637.

\bibitem{ding2019camnet}
M.~Ding, Z.~Wang, J.~Sun, J.~Shi, and P.~Luo, ``Camnet: Coarse-to-fine
  retrieval for camera re-localization,'' in \emph{Proceedings of the IEEE/CVF
  International Conference on Computer Vision}, 2019, pp. 2871--2880.

\bibitem{balntas2018relocnet}
V.~Balntas, S.~Li, and V.~Prisacariu, ``Relocnet: Continuous metric learning
  relocalisation using neural nets,'' in \emph{Proceedings of the European
  Conference on Computer Vision (ECCV)}, 2018, pp. 751--767.

\bibitem{laskar2017camera}
Z.~Laskar, I.~Melekhov, S.~Kalia, and J.~Kannala, ``Camera relocalization by
  computing pairwise relative poses using convolutional neural network,'' in
  \emph{Proceedings of the IEEE International Conference on Computer Vision
  Workshops}, 2017, pp. 929--938.

\bibitem{barnes2019masking}
D.~Barnes, R.~Weston, and I.~Posner, ``Masking by moving: Learning
  distraction-free radar odometry from pose information,'' \emph{arXiv preprint
  arXiv:1909.03752}, 2019.

\bibitem{barnes2020under}
D.~Barnes and I.~Posner, ``Under the radar: Learning to predict robust
  keypoints for odometry estimation and metric localisation in radar,'' in
  \emph{2020 IEEE International Conference on Robotics and Automation
  (ICRA)}.\hskip 1em plus 0.5em minus 0.4em\relax IEEE, 2020, pp. 9484--9490.

\bibitem{zhu2020cylindrical}
X.~Zhu, H.~Zhou, T.~Wang, F.~Hong, Y.~Ma, W.~Li, H.~Li, and D.~Lin,
  ``Cylindrical and asymmetrical 3d convolution networks for lidar
  segmentation,'' \emph{arXiv preprint arXiv:2011.10033}, 2020.

\bibitem{wang2020atloc}
B.~Wang, C.~Chen, C.~X. Lu, P.~Zhao, N.~Trigoni, and A.~Markham, ``Atloc:
  Attention guided camera localization,'' in \emph{Proceedings of the AAAI
  Conference on Artificial Intelligence}, vol.~34, no.~06, 2020, pp.
  10\,393--10\,401.

\bibitem{huang2019prior}
Z.~Huang, Y.~Xu, J.~Shi, X.~Zhou, H.~Bao, and G.~Zhang, ``Prior guided dropout
  for robust visual localization in dynamic environments,'' in
  \emph{Proceedings of the IEEE/CVF International Conference on Computer
  Vision}, 2019, pp. 2791--2800.

\bibitem{he2016deep}
K.~He, X.~Zhang, S.~Ren, and J.~Sun, ``Deep residual learning for image
  recognition,'' in \emph{Proceedings of the IEEE conference on computer vision
  and pattern recognition}, 2016, pp. 770--778.

\bibitem{locatello2020object}
F.~Locatello, D.~Weissenborn, T.~Unterthiner, A.~Mahendran, G.~Heigold,
  J.~Uszkoreit, A.~Dosovitskiy, and T.~Kipf, ``Object-centric learning with
  slot attention,'' \emph{Advances in Neural Information Processing Systems},
  vol.~33, pp. 11\,525--11\,538, 2020.

\bibitem{hong2020radarslam}
Z.~Hong, Y.~Petillot, and S.~Wang, ``Radarslam: Radar based large-scale slam in
  all weathers,'' in \emph{2020 IEEE/RSJ International Conference on
  Intelligent Robots and Systems (IROS)}.\hskip 1em plus 0.5em minus
  0.4em\relax IEEE, 2020, pp. 5164--5170.

\bibitem{wang2019deep}
Y.~Wang and J.~M. Solomon, ``Deep closest point: Learning representations for
  point cloud registration,'' in \emph{Proceedings of the IEEE/CVF
  international conference on computer vision}, 2019, pp. 3523--3532.

\bibitem{uy2018pointnetvlad}
M.~A. Uy and G.~H. Lee, ``Pointnetvlad: Deep point cloud based retrieval for
  large-scale place recognition,'' in \emph{Proceedings of the IEEE conference
  on computer vision and pattern recognition}, 2018, pp. 4470--4479.

\bibitem{RobotCarDatasetIJRR}
\BIBentryALTinterwordspacing
W.~Maddern, G.~Pascoe, C.~Linegar, and P.~Newman, ``{1 Year, 1000km: The Oxford
  RobotCar Dataset},'' \emph{The International Journal of Robotics Research
  (IJRR)}, vol.~36, no.~1, pp. 3--15, 2017. [Online]. Available:
  \url{http://dx.doi.org/10.1177/0278364916679498}
\BIBentrySTDinterwordspacing

\bibitem{scikit-learn}
F.~Pedregosa, G.~Varoquaux, A.~Gramfort, V.~Michel, B.~Thirion, O.~Grisel,
  M.~Blondel, P.~Prettenhofer, R.~Weiss, V.~Dubourg, J.~Vanderplas, A.~Passos,
  D.~Cournapeau, M.~Brucher, M.~Perrot, and E.~Duchesnay, ``Scikit-learn:
  Machine learning in {P}ython,'' \emph{Journal of Machine Learning Research},
  vol.~12, pp. 2825--2830, 2011.

\bibitem{levinson2010robust}
J.~Levinson and S.~Thrun, ``Robust vehicle localization in urban environments
  using probabilistic maps,'' in \emph{2010 IEEE international conference on
  robotics and automation}.\hskip 1em plus 0.5em minus 0.4em\relax IEEE, 2010,
  pp. 4372--4378.

\bibitem{wan2018robust}
G.~Wan, X.~Yang, R.~Cai, H.~Li, Y.~Zhou, H.~Wang, and S.~Song, ``Robust and
  precise vehicle localization based on multi-sensor fusion in diverse city
  scenes,'' in \emph{2018 IEEE international conference on robotics and
  automation (ICRA)}.\hskip 1em plus 0.5em minus 0.4em\relax IEEE, 2018, pp.
  4670--4677.

\bibitem{lu2019l3}
W.~Lu, Y.~Zhou, G.~Wan, S.~Hou, and S.~Song, ``L3-net: Towards learning based
  lidar localization for autonomous driving,'' in \emph{Proceedings of the
  IEEE/CVF Conference on Computer Vision and Pattern Recognition}, 2019, pp.
  6389--6398.

\bibitem{s2p2-2e66-23}
\BIBentryALTinterwordspacing
M.~Ibrahim, N.~Akhtar, S.~Anwar, M.~Wise, and A.~Mian, ``Perth-wa localization
  dataset in 3d point cloud maps,'' 2023. [Online]. Available:
  \url{https://dx.doi.org/10.21227/s2p2-2e66}
\BIBentrySTDinterwordspacing

\end{thebibliography}

\end{document}